\documentclass[pdflatex,sn-mathphys-num, iicol]{sn-jnl}

\usepackage{graphicx}%
\usepackage{multirow}%
\usepackage{amsmath,amssymb,amsfonts}%
\usepackage{amsthm}%
\usepackage{mathrsfs}%
\usepackage[title]{appendix}%
\usepackage{xcolor}%
\usepackage{textcomp}%
\usepackage{manyfoot}%
\usepackage{booktabs}%
\usepackage{algorithm}%
\usepackage{algorithmicx}%
\usepackage{algpseudocode}%
\usepackage{listings}%
\usepackage{lineno}

\usepackage{xspace}
\makeatletter
\DeclareRobustCommand\onedot{\futurelet\@let@token\@onedot}
\def\@onedot{\ifx\@let@token.\else.\null\fi\xspace}
\makeatother
\newcommand{\eg}{\emph{e.g.}\xspace}

\usepackage{mathrsfs} 
\usepackage{mathalfa}
\usepackage{colortbl}
\usepackage{multirow}

\usepackage{soul}
\usepackage{hyperref}
\usepackage{tabularx}
\usepackage{graphicx}
\usepackage{array}

\theoremstyle{thmstyleone}%
\theoremstyle{thmstyletwo}%

\theoremstyle{thmstylethree}%

\begin{document}

\title[Article Title]{Paint Outside the Box: Synthesizing and Selecting Training Data for Visual Grounding}


\author[1]{\fnm{Zilin} \sur{Du}}\email{zilin003@e.ntu.edu.sg}

\author[1]{\fnm{Haoxin} \sur{Li}}\email{haoxin003@e.ntu.edu.sg}

\author[2]{\fnm{Jianfei} \sur{Yu}}\email{jfyu@njust.edu.cn}

\author*[1]{\fnm{Boyang} \sur{Li}}\email{boyang.li@ntu.edu.sg}

\affil*[1]{\orgdiv{College of Computing and Data Science}, \orgname{Nanyang Technological University}, \orgaddress{\city{Singapore}, \country{Singapore}}}

\affil[2]{\orgdiv{School of Computer Science and Engineering}, \orgname{Nanjing University of Science and Technology}, \orgaddress{\city{Nanjing}, \country{China}}}



\abstract{Visual grounding aims to localize the image regions based on a textual query. Given the difficulty of large-scale data curation, we investigate how to effectively learn visual grounding under data-scarce settings in this paper. To address the data scarcity, we propose a novel framework, POBF (Paint Outside the Box and Filter). POBF synthesizes images by inpainting outside the box, tackling a label misalignment issue encountered in previous works. Furthermore, POBF leverages an innovative filtering scheme to select the most effective training data. This scheme combines a hardness score and an overfitting score, balanced by a penalty term. Extensive experiments across four benchmark datasets demonstrate that POBF consistently improves performance, achieving an average gain of 5.83\% over the real-data-only method and outperforming leading baselines by 2.29\%–3.85\% in accuracy. We validate the robustness and generalizability of POBF across various generative models, training data sizes, and model architectures.}

\keywords{Visual Grounding, Cross-Modality Data Generation, Data Augmentation, Image Editing, Data Filtering}



\maketitle

\section{Introduction}
\label{sec:intro}

Visual grounding \cite{deng2021transvg, xiao2024towards, ye2023whether, yu2024revisiting, su2024scanformer, liu2025groundingmate, cao2025flashvtg, yi2024augment, ma2024visual} is a fundamental task in vision-language understanding \cite{li2025perception, li_unimoe}. The objective is to locate a specific image region based on a free-form textual query. However, constructing visual grounding datasets with dense region-text annotations is both time-intensive and costly, which hinders the progress in this field. Motivated by the difficulty of large-scale data curation, we investigate the feasibility of effectively learning visual grounding under data-scarce (commonly referred to as few-shot \cite{xue2024few}, data-efficient \cite{chen2022visualgpt}, or low-resource \cite{jiang2023mixphm}) settings, where only limited real training data are accessible.

\begin{figure*}[t]
  \centering
   \includegraphics[width=1.0\linewidth]{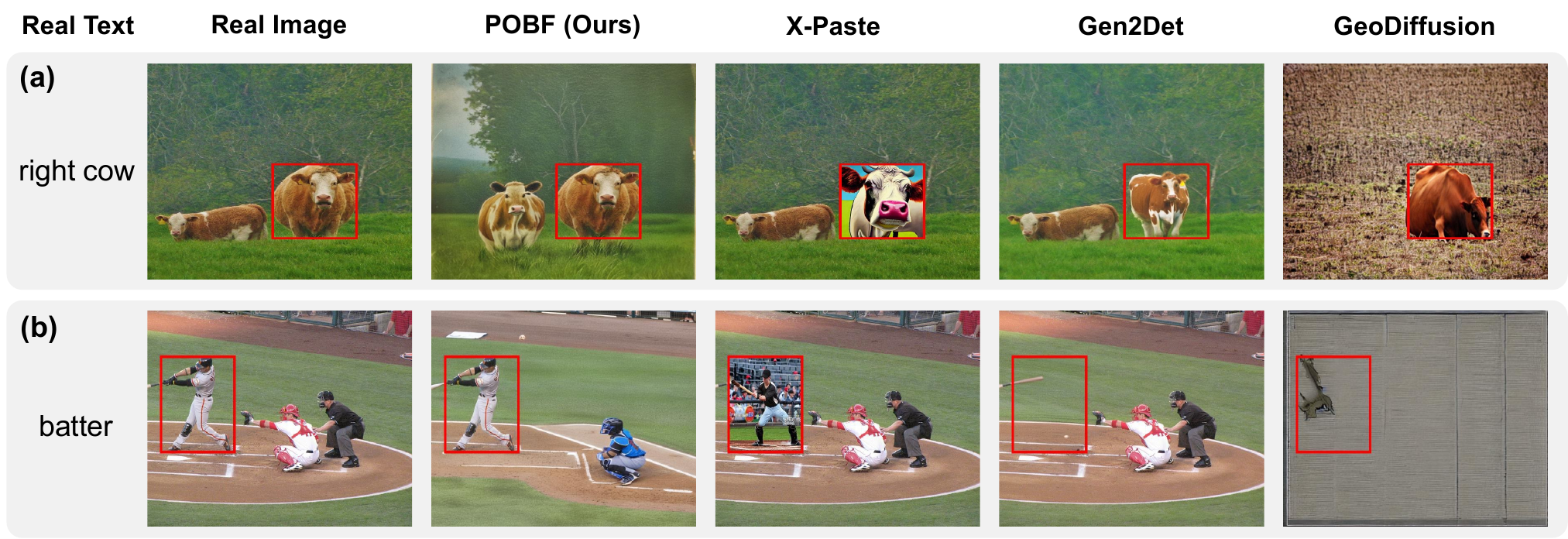}
   \caption{Examples of generated images from different methods. X-Paste \cite{zhao2023x} synthesizes unnatural images. Gen2Det \cite{suri2024gen2det} exhibits the label misalignment issue. GeoDiffusion  \cite{chengeodiffusion} either fails to generate the correct object or introduces excessive artifacts. In contrast, our method synthesizes high-quality images with accurately aligned bounding boxes.}
   \label{fig:examples}
\end{figure*}

With recent advances in image generation and captioning \cite{li2022blip,podellsdxl,rombach2022high,liu2024visual}, a natural approach to address data scarcity is to synthesize training data \cite{dunlap2023diversify, wang2024enhance}. These powerful generative models, pretrained on large-scale real-world data, have captured the correspondence between images and textual descriptions. It is probable that distilling such knowledge can benefit visual grounding through training data generation. Many recent works \cite{zhao2023x, fan2024divergen, yang2024freemask, feng2024instagen, li2023open, fang2024data, chengeodiffusion, chen2024rtgen} have explored the generation of densely annotated data to train models for fine-grained understanding tasks, such as object detection and image segmentation. However, they often rely on specialized networks trained on extensive dense annotations, which limits their flexibility and generalizability.

To overcome this limitation, we propose to leverage off-the-shelf generative models \cite{podellsdxl,li2022blip,rombach2022high}, trained solely on image-caption pairs in a weakly-supervised manner, for training data generation. Compared to the methods \cite{wang2025learning, yang2024freemask, wang2024detdiffusion, zheng2024resvg, he2024learning} relying on specialized networks, our method offers two advantages. First, our method could generalize across a wider range of domains, since the generative models we used are trained on web-scale data. Second, we can benefit from the rapid progress in generative models, incorporating the latest advancements into our data generation pipeline without additional training on dense annotations.

Although the off-the-shelf generative models are capable of synthesizing images across diverse domains, they struggle to directly generate data with region-text labels, as they are not trained with dense annotations. Therefore, we leverage the limited region-text annotations available in real data to generate training data through inpainting. However, a naive approach—inpainting the object within the bounding box (\eg, Gen2Det \cite{suri2024gen2det}, see Figure \ref{fig:examples})—can result in label misalignment issues. Specifically, the generated object may not spatially align well with the box or even fail to appear entirely. For example, the synthetic `right cow' in Figure \ref{fig:examples}(a) misaligns with the box, and the `batter' in Figure \ref{fig:examples}(b) disappears. To tackle the label misalignment problem, we propose a paint-outside-the-box strategy, which preserves the content within the box while generating the background outside. This results in higher-quality images that align well with the original annotations.

Moreover, training neural networks with synthetic data presents another challenge: not all samples contribute positively to the model performance \cite{chen2024rtgen, chen2024auto,malach2017decoupling, mahmoud2024sieve}. To identify the most effective training data, we propose a filtering scheme consisting of three components: a hardness score, an overfitting score, and a penalty term. The hardness score prioritizes the easiest samples, as it is challenging to model outliers when the basic patterns are not adequately captured in our data-scarce regime \cite{sorscher2022beyond}. The overfitting score penalizes samples exacerbating overfitting the unintended features outside the box, as shortcut features from the background could lead the gigantic model to overfit \cite{geirhos2020shortcut, beery2018recognition, xiao2020noise}, especially with limited training data. To manage the trade-off between two scores, we introduce a penalty term.

In summary, we propose a novel framework named POBF (Paint Outside the Box and Filter). 
POBF synthesizes high-quality training samples by image inpainting and captioning, then employs an effective filtering scheme for data selection. Our contributions are:
\begin{itemize} 
\item We address a challenging problem of visual grounding in data-scarce scenarios by introducing POBF, a flexible and adaptable framework that leverages off-the-shelf data generators pretrained solely on image–caption pairs.
\item We propose a novel data generation strategy for visual grounding, paint outside the box, which overcomes the label misalignment issues in previous approaches.
\item We introduce a new filtering scheme designed for data-scarce settings to identify the most effective synthetic training samples. This scheme integrates a hardness score, an overfitting score, and a penalty term.
\item Experiments demonstrate that POBF outperforms strong baselines by a 2.29\%–3.85\% margin. We further validate its robustness and generalizability across generative models, training data sizes, and model architectures.
\end{itemize}

\section{Related Work}
\label{sec:related_work}

\noindent
\textbf{Visual Grounding.} Early approaches addressed visual grounding in a two-stage framework \cite{liu2019learning, yu2018mattnet, liu2019improving}. Recently, more efficient one-stage methods have been developed \cite{deng2021transvg, kamath2021mdetr, ye2022shifting, su2023language, deng2023transvg++, xiao2023clip, shi2023dynamic, xiao2024hivg, yang2022improving, yao2024visual, chen2022multi, jiao2023suspected, weerakoon2022softskip, liang2023luna}. QRNet \cite{ye2022shifting}, VG-LAW \cite{su2023language}, and TransVG++ \cite{deng2023transvg++}, integrate language-guided knowledge within the intermediate layers of visual backbones to compute query-dependent visual attention dynamically. As the first diffusion-based framework, LG-DVG \cite{chen2023language} progressively reasons the queried object by denoising noisy boxes. 

With the development of pretrained large multimodal models, \cite{he2024improved, yang2023improving, han2024zero, kang2025your, yan2024vigor, li2025uniomni} leverage the foundation models to localize objects. For example, \cite{he2024improved, yang2023improving} employ GradCAM \cite{selvaraju2017grad} to produce gradient-based visual explanations, represented as heatmaps for grounding outputs. Inspired by reinforcement learning, ViGoR \cite{yan2024vigor} integrates a reward model for instruction tuning for visual grounding. In parallel, \cite{chen2024lion, chen2023shikra, zhao2024llm, zeng2024investigating} focus on training multimodal large language models for visual grounding using extensive, fine-grained labeled datasets.

Previous works typically train models on large-scale labeled visual grounding datasets. To mitigate the issue of data scarcity, several studies have explored unsupervised \cite{wang2024omni}, weakly-supervised \cite{chen2024querymatch, liu2021relation, sun2021discriminative, jin2023refclip, liu2023confidence, lin2024triple}, and semi-supervised \cite{sun2023refteacher} learning methods. However, these methods still rely heavily on the availability of large amount of in-domain real data. Instead, we explore a more constrained data-scarce setting, where only a limited amount of real visual grounding data exists.

\vspace{0.5em}
\noindent
\textbf{Training with Synthetic Data.} Building off image generation and image captioning, training on synthetic data has emerged as a promising research direction \cite{yang2023ai, rotstein2024fusecap, xuan2023distilling, tian2023learning, du2023training, bafghi2025mixdiff, petit2025echodfkd, kim2025pixswap, panev2024exploring, kim2024controllable, zhu2024odgen, hesynthetic}. In fine-grained understanding tasks, researchers explore the use of synthetic data in object detection \cite{fang2024data, wang2024detdiffusion, chengeodiffusion, zhang2023diffusionengine, feng2024instagen, suri2024gen2det}, image segmentation \cite{yang2024freemask, li2023open, xie2024mosaicfusion, fan2024divergen, zhao2023x}, and visual grounding \cite{he2024learning, wang2024learning, jiang2022pseudo}. \cite{li2023open, zhang2023diffusionengine, feng2024instagen} incorporate task-specific heads into image generators and fine-tune them with extensive data to synthesize images paired with annotations. In addition, \cite{zhao2023x, fan2024divergen, he2024learning} first generate images, then employ an auxiliary model to annotate the synthetic data. Furthermore, \cite{fang2024data, chengeodiffusion, yang2024freemask, wang2024detdiffusion} employ visual priors (\eg, segmentation masks or edges) to guide the synthetic image generation via controllable diffusion models. However, these methods rely on either the data generator, the annotator, or both to be pretrained on large-scale densely annotated datasets. 


Some prior works \cite{jiang2022pseudo, wang2024learning, he2024learning, zheng2024resvg, wang2025learning} have focused on generating labeled training data for visual grounding, but they typically rely on specialized networks—such as object detectors or image generators trained on extensively annotated datasets—which limits their generalizability and flexibility. In contrast, we introduce a novel training data generation method that leverages off-the-shelf generative models trained solely on image–caption pairs, eliminating the reliance on extensive dense annotations and making our approach more flexible. In addition, we incorporate a filtering scheme tailored for data-scarce visual grounding.

\begin{figure*}[t]
  \centering
   \includegraphics[width=1.0\linewidth]{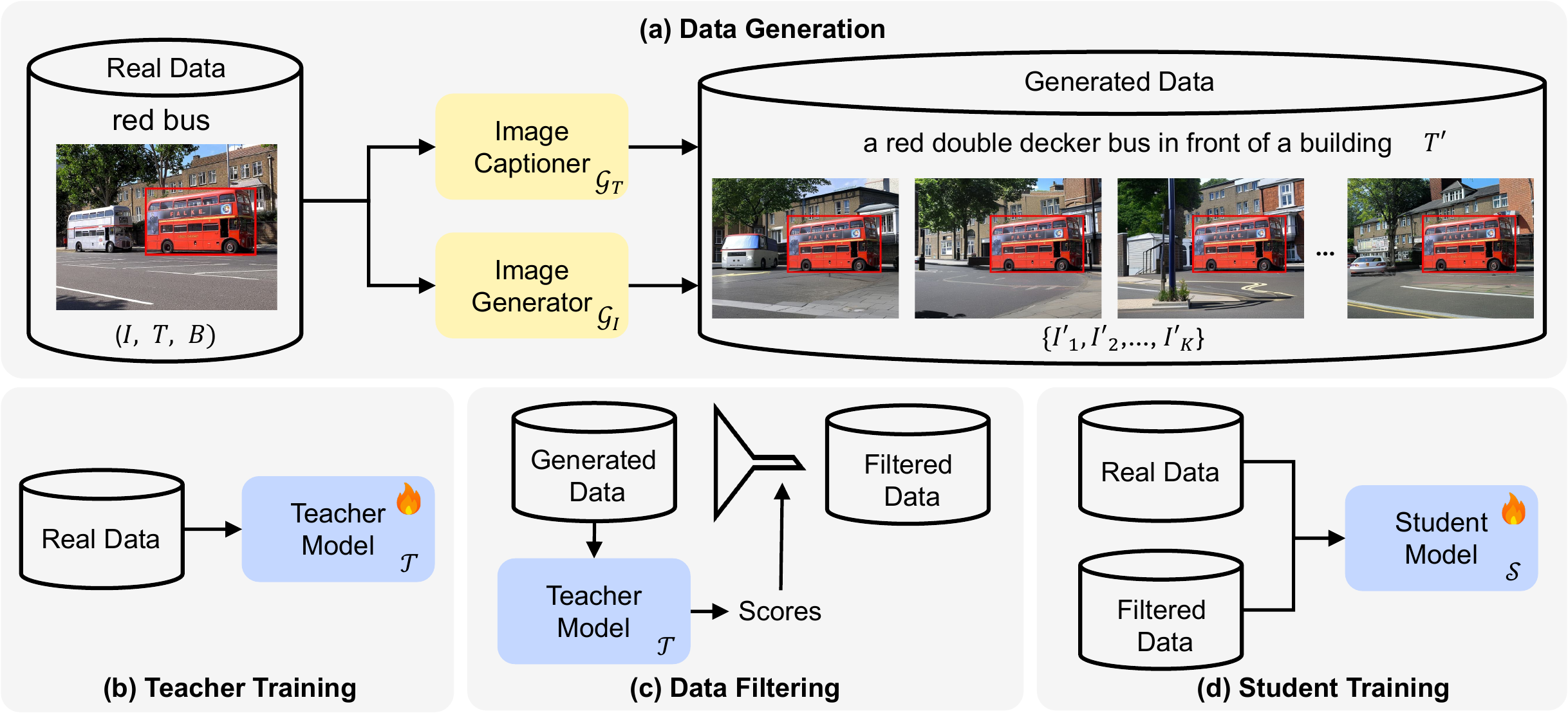}
   \caption{Overview of our proposed framework POBF, which consists of four steps: data generation, teacher training, data filtering, and student training. The real data sample (image, caption, box) is denoted as $(I, T, B)$. $T'$ and $I'_j$ denote the generated caption and image, respectively. $\mathcal{G}_{T}$ and $\mathcal{G}_{I}$ are generative models. $\mathcal{T}$ and $\mathcal{S}$ refer to the teacher and student model, respectively.}
   \label{fig:overview}
\end{figure*}

\vspace{0.5em}
\noindent
\textbf{Data Pruning.} Recent advancements in data selection methods have gained significant attention from the research community \cite{xie2023data, jiang2024adaptive, mahmoud2024sieve,guo2024generating, gupta2024training, pmlr-v306-just26a, Li_2025_CVPR, du2026scalable, pmlr-v306-du26u}. These methods typically employ pre-defined criteria to assign a scalar score to each training example. While simple heuristics like repetition \cite{lee2021deduplicating}, string existence \cite{penedo2023refinedweb}, or semantic similarity \cite{abbas2023semdedup} are commonly used, more sophisticated criteria have also been explored, including gradient norms \cite{paul2021deep}, forgetfulness \cite{toneva2018empirical}, and influence function scores \cite{yang2022dataset}. A common strategy is to prioritize data samples with lower training loss \cite{yao2020searching,chen2019understanding}. This approach leverages the observation that DNNs initially learn generalizable patterns before gradually overfitting to noisy data \cite{song2019does, li2020gradient}. To counter confirmation bias, techniques such as co-training models \cite{yu2019does, li2020dividemix} have been proposed, where two networks are trained concurrently and use their combined predictions for selection. MentorNet \cite{MentorNet-jiang18c} further extends this idea by introducing a learned curriculum to guide the selection of training data. 

Our use of separately trained teacher networks contrasts with recent studies \cite{chen2024rtgen, rose2023visual, yang2022z, tang2023learning, he2024learning} on synthetic training data selection, which filter out text-image pairs with low CLIP similarity scores. In contrast to prior work, we propose to identify the most effective synthetic training data from the hardness and overfitting perspectives.

\section{Problem Definition}
\label{sec:problem_definition}

Visual grounding is the task of identifying the target region \( R \subseteq I \) within the image that a natural language \( T \) refers to. Typically, \( R \) is represented as a bounding box \( B = (x, y, w, h) \), where \( (x, y) \) specifies the box center coordinates, and \( w \) and \( h \) define its width and height. In this paper, we investigate the problem of learning visual grounding in data-scarce scenarios, namely data-scarce visual grounding, where only a very limited amount of annotated real training data is available. Following \cite{xue2024few, chen2022visualgpt, jiang2023mixphm}, we simulate this setting by randomly sampling a small subset of the full training set. 

\section{Methodology}
\label{sec:methodology}

To address the data-scarce visual grounding problem, we introduce a novel, model-agnostic framework called POBF (Paint Outside the Box and Filter). As illustrated in Figure~\ref{fig:overview}, POBF consists of four stages. (a) Our approach begins with a data generation process, where we synthesize images and captions using image inpainting and captioning techniques. This process utilizes the image generator and captioner pretrained on large-scale image-caption pairs. (b) Next, we train a teacher model using the limited real data. (c) We then apply a filtering scheme to select effective synthetic images. Specifically, the well-trained teacher network assigns a hardness score, an overfitting score, and a penalty term to each synthetic image. (d) Finally, we retain the synthetic images with high scores and combine them with real data to train the student model. The proposed POBF framework is simple, effective, and does not rely on any auxiliary models pretrained with dense annotated data.

\subsection{Cross-modality Data Generation}

\noindent \textbf{Image Generation with Inpainting.} We begin by synthesizing images through image inpainting, a process that regenerate masked regions within an image based on a given prompt, while keeping the unmasked areas intact. To address the label misalignment problem, we propose to paint outside of the bounding box instead. For each data point (\( I, T, B \)), we first use a captioning model \( \mathcal{G}_{T} \) to generate a caption \( C \) for \( I \). The bounding box \( B \), image \( I \), and caption \( C \) are then input into the inpainting model \( \mathcal{G}_{I} \), yielding a generated image \( I' \). As shown in Figure~\ref{fig:overview}(a), the red bus within the bounding box remains unchanged, making it strictly align with the ground truth bounding box after generation. The synthetic image \( I' \) can be paired with the original text \( T \) and bounding box \( B \) to form a new training sample (\( I', T, B \)). By generating \( K \) synthetic images per real sample, denoted as $\{I'_1, I'_2, \dots, I'_K\}$, the training dataset effectively increases in size by a factor of \( K \). A key hyperparameter in the inpainting process is \emph{strength}, which controls the amount of noise added to the base image and, consequently influences the similarity between the generated output and the original image. To encourage data diversity, we set the \emph{strength} parameter to a high value.

\vspace{0.5em}
\noindent \textbf{Text Generation with Captioning.} To augment the original text \( T \), we feed the cropped image within the bounding box \( B \) to the captioning model \( \mathcal{G}_{\text{T}} \), which generates a new caption \( T' \) describing the object. The generated caption \( T' \) can be paired with $B$ and either the real image \( I \) or the generated image \( I' \) to create a new training sample. In this work, we employ $p$-sampling \cite{holtzman2019curious} to improve diversity.

\subsection{Filtering Scheme}

\begin{figure}[t]
  \centering
   \includegraphics[width=\linewidth]{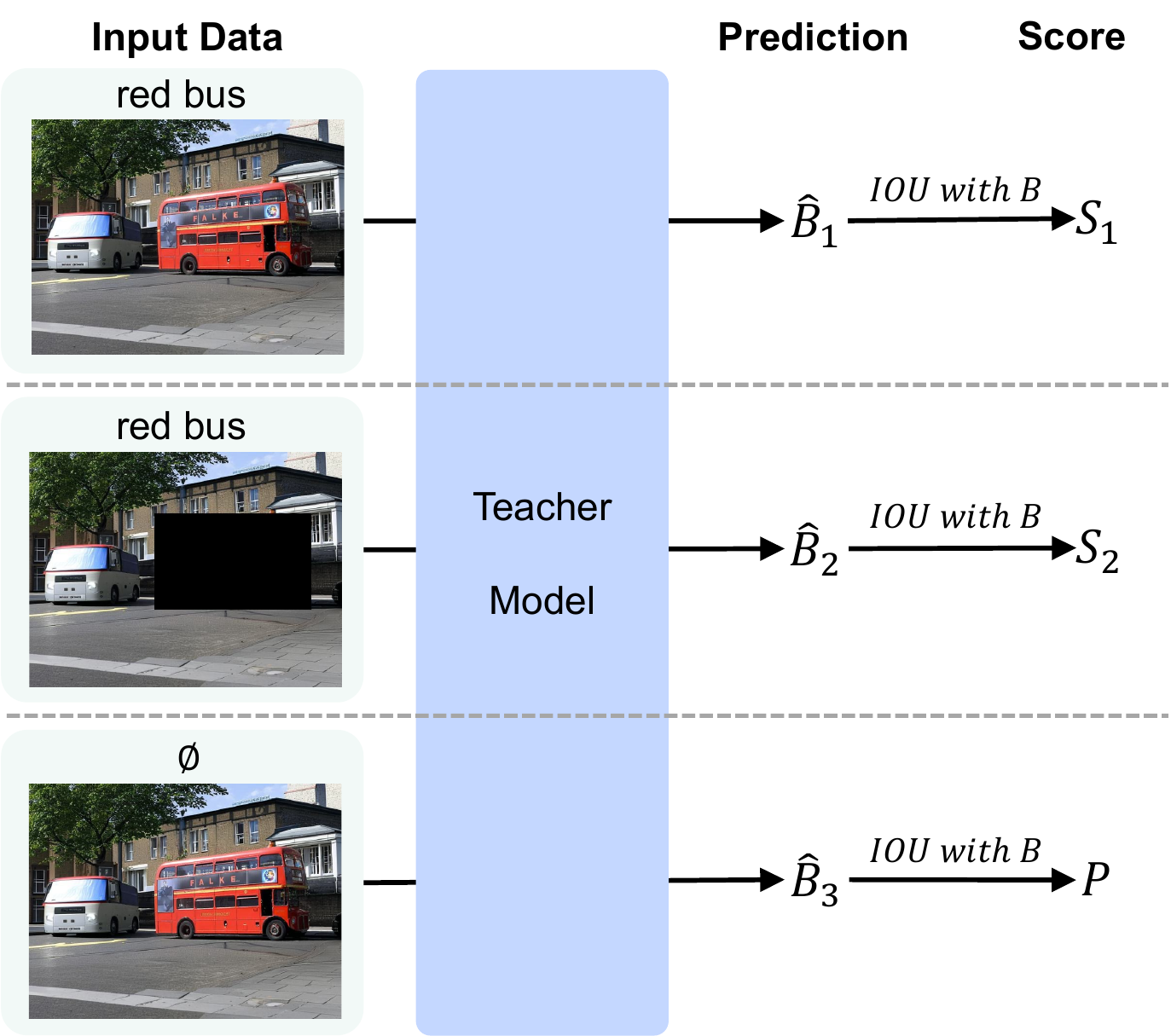}
   \caption{Illustration of the inputs used to compute each score for a generated image. Here, $B$ and $\hat{B}$ denote the ground truth and predicted bounding box, respectively.}
   \label{fig:score}
\end{figure}

Recognizing that not all synthetic samples contribute positively to model performance \cite{chen2024rtgen, chen2024auto,malach2017decoupling, mahmoud2024sieve}, we propose a novel filtering scheme to select the most effective images for training. Initially, we use the limited real training dataset to train a teacher model \( \mathcal{T} \). Then, \( \mathcal{T} \) evaluates synthetic images based on two scores and one penalty term to determine their effectiveness as training samples. We provide an illustration in Figure~\ref{fig:score} to demonstrate the input used when calculating each score for a synthetic image. For clarity in the following discussions, we assume a real data sample \((I, T, B)\) and denote the set of images generated from it as $\{I'_1, I'_2, \dots, I'_K\}$.

\vspace{0.5em}
\noindent \textbf{Hardness Score.} We propose to filter harmful synthetic images via computing hardness. Intuitively, in our data-scarce regime, it is difficult to model outliers effectively as fundamental patterns may not be fully captured \cite{sorscher2022beyond}. Since the model requires easiest examples to provide coarse-grained information about the target function, we prioritize synthetic images that are less challenging for the teacher model \( \mathcal{T} \). We define the hardness score for each of the synthetic image \( I_j' \), where $j \in \{1, 2, \dots, K\}$, as follows:
\begin{equation}
    S_1(I_j') = \text{IOU}(\mathcal{T}(I_j', T), B)
\end{equation}
where $\mathcal{T}(I_j', T)$ is the predicted bounding box, and IOU (Intersection over Union) is the ratio of the intersection of two areas to their combined area.


\vspace{0.5em}
\noindent \textbf{Overfitting Score.} Additionally, we introduce another score to filter out synthetic images that may hinder training by exacerbating overfitting. Given the gigantic model architecture relative to the limited number of real data, the model is prone to overfit by relying on image background features rather than the actual object in the bounding box \cite{beery2018recognition, xiao2020noise}. To mitigate this, we assess the potential of each generated image to contribute to overfitting. Specifically, we modify each generated image by masking the area within the bounding box and input this variant into \( \mathcal{T} \). If \( \mathcal{T} \) can still correctly predict the bounding box, it indicates that \( \mathcal{T} \) may leverage unintended features present in the generated background. Incorporating such data into training dataset risks the model learning these unintended features, thereby exacerbating overfitting. We define an overfitting score to filter these problematic samples. The score for each generated image \( I_j' \), where $j \in \{1, 2, \dots, K\}$, is defined as:
\begin{equation}
    S_2(I_j') = 1 - \text{IOU}(\mathcal{T}(\text{Mask}(I_j', B), T), B)
\end{equation}
where \( \text{Mask}(I_j', B) \) is a operation that sets all pixel values within the bounding box \( B \) to zero.

\vspace{0.5em}
\noindent \textbf{Penalty Term.} To balance between preserving easy samples and the mitigation of overfitting risk, we introduce a penalty term. The rationale is that overly easy images may not challenge the model enough to learn complex patterns. Conversely, images where the background outside the bounding box does not offer any useful information may introduce model bias. To balance between the first two scores, we define a penalty term for each generate image \( I_j' \), where $j \in \{1, 2, \dots, K\}$, as follows:
\begin{equation}
    P(I_j') = \text{IOU}(\mathcal{T}(I_j', \emptyset),B)
\end{equation}
where $\emptyset$ denotes an empty textual string. 
This term can be regarded as measuring the target-related visual prior in the generated image. Specifically, it measures how strongly the generated image alone supports the ground-truth box, independent of the referring expression. A high value of \(P\) indicates that the target remains visually identifiable from language-independent visual evidence. Together with \(S_2\), \(P\) acts as a regularizing term by penalizing synthetic samples whose target region contains insufficient visual evidence, through assigning them a lower overall score. In this way, the selected samples are encouraged to retain strong target-related visual cues while avoiding excessive reliance on the synthesized background.

The three scores are individually normalized to have zero mean and unit standard deviation. Then The final score for a generated image \( I_j' \) is calculated as:
\begin{equation}
    \text{Score}(I_j') = \lambda_1 S_1(I_j') + \lambda_2 S_2(I_j')+ \lambda_P P(I_j')
\end{equation}
where $\lambda_1$, $\lambda_2$, and $\lambda_P$ are hyperparameters controlling the weights of each component.

Out of the \( K \) generated images \( \{I'_1,I'_2,\dots,I'_K\} \) from each real data point (\( I, T, B \)), we retain one synthetic image \( I'_m \) with the highest score, where \( m = \arg\max_j \text{Score}(I'_j) \) and \( j \in \{1, 2, \dots, K\} \). This yields a synthetic data point (\( I', T, B \)). By incorporating the selected synthetic data points into the training set, we effectively double its size. Finally, we train a student model $\mathcal{S}$, where the real text \( T \) is randomly replaced with the corresponding generated caption \( T' \) with probability \( q \) during training.

\section{Experiments}
\label{sec:experiments}

\noindent\textbf{Dataset.} 
We conduct experiments on RefCOCO \cite{yu2016modeling}, RefCOCO+ \cite{yu2016modeling}, RefCOCOg \cite{mao2016generation, nagaraja2016modeling}, and ReferItGame \cite{kazemzadeh2014referitgame}. Dataset statistics are in Table \ref{tab:data_statistics}. We use top-1 accuracy as the evaluation metric. For each dataset, we randomly sample subsets comprising 1\% of the entire dataset to simulate data-scarce scenarios \cite{xue2024few, chen2022visualgpt, jiang2023mixphm}.

\begin{table}[h!]
\centering
\begin{tabular}{c c c c}
\hline
\textbf{Dataset} & \textbf{Images} & \textbf{Texts} & \textbf{Objects} \\ \hline
RefCOCO \cite{yu2016modeling} & 19,994   & 142,210 & 50,000  \\ 
RefCOCO+ \cite{yu2016modeling}  & 19,992   & 141,564 & 49,856  \\
RefCOCOg \cite{mao2016generation, nagaraja2016modeling}  & 25,799   & 95,010  & 49,822  \\
ReferItGame \cite{kazemzadeh2014referitgame} & 19,894   & 130,525 & 96,654  \\
 \hline
\end{tabular}
\caption{Statistics of visual grounding datasets.}
\label{tab:data_statistics}
\end{table}

\begin{table*}
\centering
\small
\caption{Performance comparison with leading baselines under the data-scarce visual grounding setting. All methods employ the proposed filtering scheme.}
\label{tab:tab_main_result}
\resizebox{\linewidth}{!}{
\begin{tabular}{c|cc|cc|cc|c|c|c}
    \toprule
    \multirow{2}*{Method}& \multicolumn{2}{c|}{\textbf{Dataset}} & \multicolumn{2}{c|}{\textbf{RefCOCO}} & \multicolumn{2}{c|}{\textbf{RefCOCO+}} & \textbf{RefCOCOg} & \textbf{ReferIt} & \multirow{2}*{Avg} \\
    & Real & Syn & TestA & TestB & TestA & TestB & Test & Test & \\
    \midrule
    \multicolumn{10}{c}{\textit{Results on One Sampled Dataset}} \\
    Real & \checkmark & & 44.26 & 34.34 & 23.58 & 17.70 & 24.87 & 21.87 & 27.78 \\
    GeoDiffusion \cite{chengeodiffusion} & \checkmark & \checkmark &47.22 & 34.90 & 28.01 & 18.83 & 25.39 & 24.20 & 29.76\textsubscript{\textcolor{green}{+1.98}} \\
    Gen2Det \cite{suri2024gen2det} &\checkmark & \checkmark & 46.61 & 35.19 & \underline{28.85} & \underline{20.65} & 26.10 & 25.00 & 30.40\textsubscript{\textcolor{green}{+2.62}} \\
    X-Paste \cite{zhao2023x} & \checkmark & \checkmark & \underline{47.85} & \underline{39.01} & 27.59 &\textbf{21.25} & \underline{27.08} & \underline{25.16} & \underline{31.32\textsubscript{\textcolor{green}{+3.54}}}\\
    POBF (Ours) & \checkmark & \checkmark & \textbf{52.77} & \textbf{40.96} & \textbf{31.37} & \textbf{21.25} & \textbf{29.98} & \textbf{25.28} & \textbf{33.61\textsubscript{\textcolor{green}{+5.83}}} \\
    \midrule
    \multicolumn{10}{c}{\textit{Average Results across Three Sampled Datasets}} \\
    Real & \checkmark &  & 43.64\textsuperscript{$\pm$0.98} & 34.89\textsuperscript{$\pm$0.50} & 25.15\textsuperscript{$\pm$1.43} & 18.02\textsuperscript{$\pm$1.48} & 24.92\textsuperscript{$\pm$1.05}& 22.94\textsuperscript{$\pm$0.97} & 28.26 \\
    POBF (Ours) & \checkmark & \checkmark & \textbf{51.25}\textsuperscript{$\pm$1.49} & \textbf{39.62}\textsuperscript{$\pm$1.16} & \textbf{32.15}\textsuperscript{$\pm$0.68} & \textbf{22.38}\textsuperscript{$\pm$1.01} & \textbf{30.73}\textsuperscript{$\pm$0.69} & \textbf{25.10}\textsuperscript{$\pm$0.30} & \textbf{33.54\textsubscript{\textcolor{green}{+5.28}}} \\
    \bottomrule
\end{tabular}}
\end{table*}

\noindent
\textbf{Baselines.} We compare our method with three leading baselines under the data-scarce visual grounding setting. (i) X-Paste \cite{zhao2023x} uses an ensemble of pretrained segmentation models to identify instances from generated images, which are then copied into real images. (ii) Gen2Det \cite{suri2024gen2det} employs a grounded editing diffusion model to edit objects in real images. (iii) GeoDiffusion \cite{chengeodiffusion} utilizes geometric control via text prompts to enhance text-to-image diffusion models for direct object detection data generation. For X-Paste and Gen2Det, we use adapted variants that do not rely on models pretrained on large-scale densely annotated datasets, enabling a more controlled comparison with POBF. For GeoDiffusion, we follow its standard configuration, which uses a model pretrained on COCO-Stuff annotations. Implementation details are in the supplementary material.

\vspace{0.5em}
\noindent
\textbf{Evaluation Metric.} Following previous works \cite{deng2021transvg, deng2023transvg++}, we use top-1 accuracy as the evaluation metric, where a prediction is considered correct if the IoU between the predicted bounding box and the ground-truth exceeds 0.5.

\vspace{0.5em}
\noindent
\textbf{Implementation Details.} For both the teacher and student networks, we utilize TransVG \cite{deng2021transvg}, one of the most popular models for visual grounding without any bells or whistles. We utilize the Stable Diffusion XL model \cite{podellsdxl} for image inpainting, with 45 denoising steps, a strength of 0.9, and a guidance scale of 7.5. BLIP ViT-L \cite{li2022blip} is employed as the image captioner, with nucleus sampling ($p$=0.9). The hyperparameters $\lambda$ in filtering scheme is tuned using grid search on the validation set, with values of 1.0 or 0.5. The selected $(\lambda_1,\lambda_2,\lambda_P)$ values are $(0.5,0.5,1.0)$ for RefCOCO and ReferIt, and $(1.0,0.5,0.5)$ for RefCOCO+ and RefCOCOg. Each of score in the filtering scheme is independently standardized across the entire dataset. For each real data point, we generate 4 synthetic images and retain one with highest score. During training, generated texts replace real captions with a probability of $q=0.3$. Following \cite{zhang2023diffusionengine, wang2024detdiffusion, yang2024freemask}, we adopt a consistent training schedule (batch size × steps) across all experiments.

\subsection{Main Results}


\noindent
\textbf{Comparison with Leading Baselines under the Data-scarce Visual Grounding Setting.} We have following observations from Table \ref{tab:tab_main_result}. \textbf{(1) Synthetic data provides substantial performance gains.} Models utilizing synthetic data generated by various baselines consistently outperform the method trained solely on real data, with at least a 1.98\% improvement. This demonstrates the strong potential of synthetic data in boosting model performance. Our approach further improves upon this, achieving an average performance gain of 5.83\% over the real-data-only baseline, which highlights the effectiveness of our generation strategy. \textbf{(2) POBF outperforms the baseline that directly generates training data by a significant margin.} GeoDiffusion is capable of directly generating images from scratch. However, it fails to correctly generate an image for \texttt{batter} in Figure~\ref{fig:examples}. This is because fine-tuning over a fixed set of objects within a specific domain limits its generalizability to novel targets. Consequently, our method surpasses GeoDiffusion by an average margin of 3.85\%. \textbf{(3) POBF consistently outperforms image editing baselines.} Similar to POBF, both Gen2Det and X-Paste generate training samples by editing real images. The strongest baseline, X-Paste, achieves an average score of 31.32\%, while our method exceeds this by 2.29\%. In Figure~\ref{fig:examples}, X-Paste produces unnatural images, and Gen2Det suffers from label misalignment issues. These quality limitations hinder model performance. In contrast, POBF generates high-quality, natural images with accurately aligned bounding boxes. In addition, we hypothesize that the relatively modest improvement on ReferIt stems from the small objects in this dataset. The inpainting process requires generating very large background regions, which may introduce visible artifacts that hinder model performance. In conclusion, POBF consistently outperforms all four strong data-generation baselines under the data-scarce visual grounding setting.

\vspace{0.5em}
\noindent
\textbf{Average Results on Three Randomly Sampled Datasets.} Additionally, we sampled three different subsets from each dataset to further evaluate performance in Table \ref{tab:tab_main_result}. While there is some performance variance due to the relatively small size of the sampled datasets, our method consistently outperforms the model trained solely on real data across four datasets. These results highlight the robustness of our approach to varying data distributions.

\begin{table*}
\centering
\small
\caption{Ablations on the synthetic data and filtering scheme. $\rm Syn_{img}$ and $\rm Syn_{txt}$ denote synthetic images and texts, respectively.}
\label{tab:ablation}
\resizebox{1.0\linewidth}{!}{
\begin{tabular}{ccc|ccc|cc|cc|c|c|c}
    \toprule
    \multicolumn{3}{c|}{\textbf{Training Data}} & \multicolumn{3}{c|}{\textbf{Filtering}}& \multicolumn{2}{c|}{\textbf{RefCOCO}} & \multicolumn{2}{c|}{\textbf{RefCOCO+}} & \textbf{RefCOCOg} & \textbf{ReferIt} & \multirow{2}*{Avg} \\
    
    Real & $\rm Syn_{img}$ & $\rm Syn_{txt}$ & $S_1$ & $S_2$ & $P$ & TestA & TestB & TestA & TestB & Test & Test & \\
    \midrule
    
    \multicolumn{13}{c}{\textit{the Effectiveness of the Synthetic Data}} \\
    \checkmark & &  & & & & 44.26 & 34.34 & 23.58 & 17.70 & 24.87 & 21.87 & 27.78 \\
    \checkmark & \checkmark & & & & & 49.97 & 39.16 & 29.01 & 20.56 & 26.67 & 22.49 & 31.31\textsubscript{\textcolor{green}{+3.53}} \\
    \checkmark & & \checkmark & & & & 47.00 & 36.19 & 25.36 & 18.74 & 26.33 & 22.13 & 29.29\textsubscript{\textcolor{green}{+1.51}}\\
    \midrule
    \multicolumn{13}{c}{\textit{the Effectiveness of the Filtering Scheme}} \\
    \checkmark & \checkmark & \checkmark & & & & 50.13 & 38.86 & 29.10 & 20.10 & 28.31 & 23.39 & 31.65\textsubscript{\textcolor{green}{+3.87}} \\
    \checkmark & \checkmark & \checkmark & \checkmark & & & 50.66 & 39.56 & 30.63 & \underline{21.76} & \textbf{30.18} & 24.50 & 32.88\textsubscript{\textcolor{green}{+5.10}} \\ 
    \checkmark & \checkmark & \checkmark & & \checkmark & & 50.29 & 39.67 & 29.95 & 21.17 & 28.42 & 24.67 & 32.36\textsubscript{\textcolor{green}{+4.58}} \\ 
    \checkmark & \checkmark & \checkmark & & & \checkmark & 51.15 & 39.24 & 30.22 & 21.44 & 29.70 & \underline{25.01} & 32.79\textsubscript{\textcolor{green}{+5.01}} \\ 
    \checkmark & \checkmark & \checkmark & \checkmark & \checkmark & & \underline{51.94} & \underline{39.75} & \underline{30.65} & \textbf{22.44} & 29.41 & 24.30 & \underline{33.08\textsubscript{\textcolor{green}{+5.30}}} \\ 
    \checkmark & \checkmark & \checkmark & \checkmark & \checkmark & \checkmark & \textbf{52.77} & \textbf{40.96} & \textbf{31.37} & 21.25 & \underline{29.98} & \textbf{25.28} & \textbf{33.61\textsubscript{\textcolor{green}{+5.83}}} \\
    
    \bottomrule
\end{tabular}}
\end{table*}

\subsection{Ablation Study}

\noindent
\textbf{Ablation on Synthetic Data.} We begin by evaluating the effectiveness of synthetic data by separately using generated images and texts, without the filtering scheme. In Table \ref{tab:ablation}, we observe that using only real data results in relatively low performance, with the model achieving an average accuracy of 27.78\%. Introducing synthetic images leads to a significant improvement, boosting performance by 3.53\% on average, demonstrating the effectiveness of our proposed inpainting strategy. When synthetic texts are applied, the improvement is modest at 1.51\%, demonstrating the importance of generated texts for performance gains.

\vspace{0.5em}
\noindent
\textbf{Ablation on Filtering Scheme.} Next, we assess the effectiveness of different components of our filtering scheme. As shown in Table \ref{tab:ablation}, the individual components—$S_1$, $S_2$, and $P$—result in performance gains of 1.23\%, 0.71\%, and 1.14\%, respectively, compared to the variant without any filtering. These results demonstrate the effectiveness of each component in selecting valuable training data. Notably, the variant excludes only the penalty term achieves an accuracy of 33.08\%, surpassing all configurations that utilize a single filtering component. The full filtering scheme yields the highest accuracy across all datasets, reaching an average accuracy of 33.61\%. This represents a substantial improvement of 1.96\% over the baseline without filtering, further underscoring the effectiveness of our proposed filter scheme. In conclusion, both synthetic data and filtering schemes play crucial roles in enhancing model performance, with the combination of these elements yielding the best results.

\subsection{Filtering Scheme Analysis}

We analyze each score in the filtering process, excluding the influence of generated text. Inspired by recent advances in data selection \cite{xia2022moderate, sorscher2022beyond, chen2024rtgen, rose2023visual, yang2022z, tang2023learning, he2024learning}, we adapt several existing methods to fit within our teacher-based selection framework. Specifically, (i) \textit{Random teacher}: randomly selects synthetic images as training data. (ii) \textit{No teacher}: uses all synthetic data indiscriminately for training. (iii) \textit{CLIP teacher}: a commonly used approach in prior works \cite{chen2024rtgen, rose2023visual, yang2022z, tang2023learning, he2024learning}, which selects synthetic images with the highest cosine similarity with the real text. (iv) \textit{Moderate-Loss} \cite{xia2022moderate}: defines the score as the difference between a sample’s loss and the class mean loss, prioritizing samples with scores near the median. (v) \textit{Moderate-DS} \cite{xia2022moderate}: replaces the score in \textit{Moderate-Loss} with the Euclidean distance between a sample’s embedding and the class center embedding, again favoring samples near the median. (vi) \textit{Difficult-Loss} \cite{sorscher2022beyond}: prioritizes hard examples with higher loss values. In addition, we include two variants of our method that select training samples based on the negative hardness score ($-S_1$) and negative overfitting score ($-S_2$), respectively.


\begin{table*}
\centering
\caption{Results obtained by isolating scores without generated text. {\colorbox{gray!20}{Results}} with a gray background indicate performance lower than the \textit{Random teacher} baseline.}
\label{tab:score_result}
\resizebox{0.55\linewidth}{!}{
\begin{tabular}{c | c c | c c | c}
    \toprule
    \multirow{2}*{Filter} & \multicolumn{2}{c|}{\textbf{RefCOCO}} & \multicolumn{2}{c|}{\textbf{RefCOCO+}} & \multirow{2}*{Avg} \\
    & TestA & TestB & TestA & TestB &  \\
    \midrule
    Random teacher & 48.52 & 38.84 & 27.47 & 20.27 & 33.76 \\
    No teacher & \cellcolor[gray]{0.9}{48.45} & \cellcolor[gray]{0.9}{38.68} & 28.40 & 20.29 & 33.96\textsubscript{\textcolor{green}{+0.20}} \\
    CLIP teacher \cite{chen2024rtgen} & 48.84 & 39.75 & 28.07 & 20.35 & 34.25\textsubscript{\textcolor{green}{+0.49}} \\
    Moderate-Loss \cite{xia2022moderate} & 48.86 & 39.17 & \cellcolor[gray]{0.9}{27.01} & \textbf{21.17} & 34.05\textsubscript{\textcolor{green}{+0.29}} \\
    Moderate-DS \cite{xia2022moderate} & 49.61 & 40.47 & 28.21 & \cellcolor[gray]{0.9}{19.84} & 34.31\textsubscript{\textcolor{green}{+0.55}} \\
    Difficult-Loss \cite{sorscher2022beyond} & \cellcolor[gray]{0.9}{46.95} & \cellcolor[gray]{0.9}{37.03} & 27.73 & \cellcolor[gray]{0.9}{19.45} & \cellcolor[gray]{0.9}{32.97\textsubscript{\textcolor{red}{-0.79}}} \\
    \midrule
    \multicolumn{6}{c}{\textit{Our Proposed Filtering Scheme}} \\
    -$S_1$& \cellcolor[gray]{0.9}{48.17} & 38.90 & \cellcolor[gray]{0.9}{25.98} & \cellcolor[gray]{0.9}{19.25} &  \cellcolor[gray]{0.9}{33.38\textsubscript{\textcolor{red}{-0.38}}} \\
    -$S_2$ & \cellcolor[gray]{0.9}{48.22} & \cellcolor[gray]{0.9}{38.51} & \cellcolor[gray]{0.9}{26.86} & \cellcolor[gray]{0.9}{19.60} & \cellcolor[gray]{0.9}{33.30\textsubscript{\textcolor{red}{-0.46}}} \\
    $S_1$ & 49.94 & \underline{40.14} & \underline{29.53} & \underline{20.67} & \underline{35.07\textsubscript{\textcolor{green}{+1.31}}} \\
    $S_2$ & 49.48 & 39.83 & 27.68 & 20.35 & 34.34\textsubscript{\textcolor{green}{+0.58}} \\
    $P$ & \underline{50.66} & 39.47 & 28.50 & 20.41 & 34.67\textsubscript{\textcolor{green}{+0.91}} \\
    $S_1$+$S_2$+$P$ & \textbf{50.98} & \textbf{40.53} & \textbf{30.25} & 20.51 & \textbf{35.56\textsubscript{\textcolor{green}{+1.81}}} \\
    \bottomrule
\end{tabular}}
\end{table*}

We have following observations in Table \ref{tab:score_result}. \textbf{(1) Simply increasing the amount of synthetic data does not guarantee improved performance.} Although the \textit{No teacher} method utilizes more data than the \textit{Random teacher} baseline, it sometimes yields inferior results. For example, it shows performance drops of 0.07\% and 0.16\% on RefCOCO. It underscores the importance of filtering out low-quality data to ensure the effectiveness of the training data. \textbf{(2) Incorporating any individual score from our filtering scheme leads to significant performance improvements over all baselines.} The three methods with any individual score outperform all baselines with gains ranging from 0.58\% to 1.31\%. Notably, they also surpass the \textit{CLIP teacher} baseline, which is commonly used for synthetic data selection. This demonstrates the effectiveness of filtering based on hardness and overfitting perspectives compared to CLIP similarity. The observation of their effectiveness aligns with our observations in Table \ref{tab:ablation}, when synthetic texts are incorporated. \textbf{(3) Combining all three scores yields the best overall performance.} Our full filtering scheme, which integrates all three components, outperforms all baselines with improvements ranging from 1.26\% to 2.60\%. In particular, it exceeds the \textit{CLIP teacher } baseline by 1.32\% on average. This result underscores the complementary nature of the three scores and their ability to be seamlessly integrated for enhanced performance. \textbf{(4) Interestingly, filtering based on negative values of $S_1$ or $S_2$ results in worse performance than the \textit{Random teacher} baseline.} For example, applying $-S_1$ and $-S_2$ leads to average performance drops of 0.38\% and 0.46\% across the two datasets. The observed performance declines when using negative scores suggests that our filtering scheme is effective in identifying valuable training data, as the introduction of poorly selected data (as indicated by negative scores) degrades model performance.

\begin{figure}[t]
  \centering
   \includegraphics[width=1.0\linewidth]{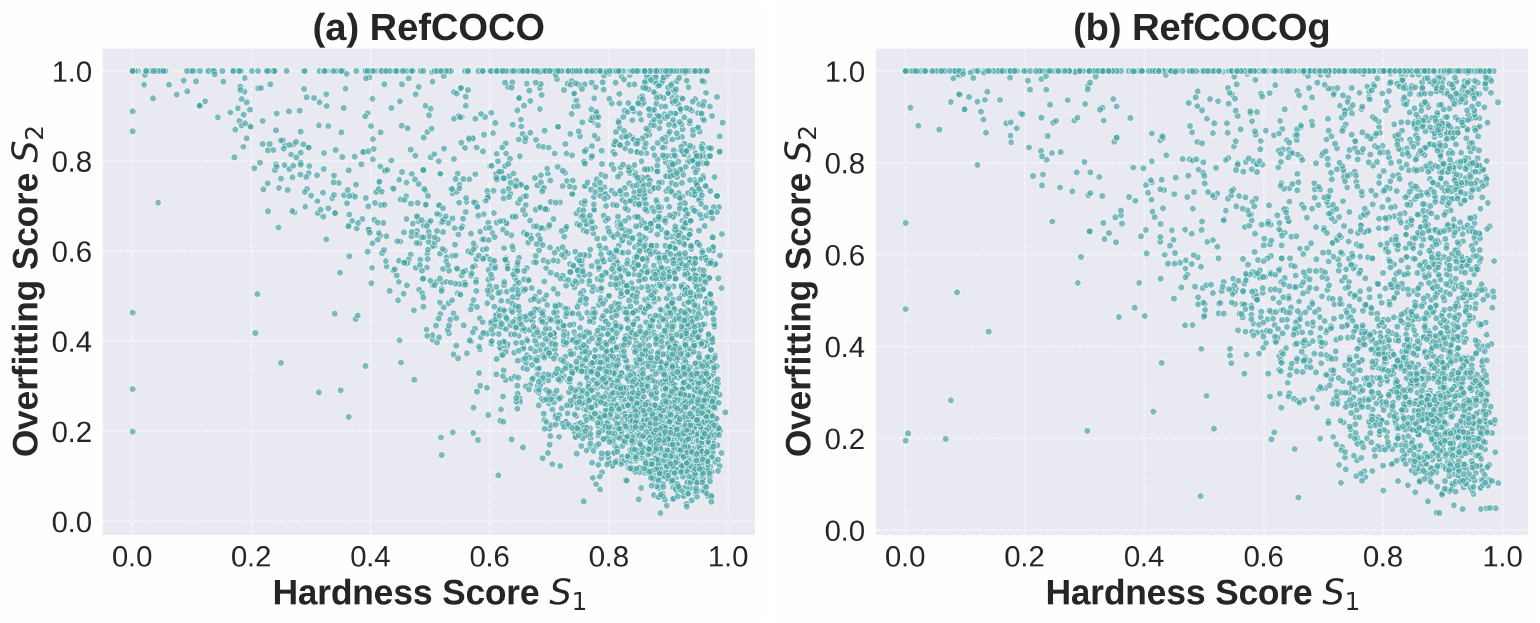}
   \caption{Scatter plot illustrating the relationship between unnormalized hardness score $S_1$ and overfitting score $S_2$ across different datasets. Each point presents a generated sample.}
   \label{fig:2Scores}
\end{figure}

\vspace{0.5em}
\noindent
\textbf{Relationship between 2 Scores.} We visualize the relationship between the hardness score $S_1$ and the overfitting score $S_2$ on the (a) RefCOCO and (b) RefCOCOg datasets in Figure \ref{fig:2Scores}. As shown, there is a weak negative correlation between the two scores. The absence of a strong linear relationship between the two scores suggests that $S_1$ and $S_2$ capture different aspects of data quality, offering complementary perspectives when selecting effective training samples. To further quantify this relationship, we compute the Pearson correlation coefficients for the two datasets, which are -0.4589 for RefCOCO and -0.3810 for RefCOCOg, respectively. These moderate correlation values highlight their potential in jointly guiding data selection.

\vspace{0.5em}
\noindent
\textbf{Experiments regards weights in filtering scheme} We tune the filtering weights via grid search, using values of 1.0 and 0.5. Table \ref{tab:hyperparameter_results} shows performance of different weights configuration across two datasets, where top-5 means the best 5 configurations on the validation datasets. Our method consistently outperforms the Real baseline by 4.1\%–5.5\%, indicating low sensitivity to weight choices and demonstrating both effectiveness and robustness.

\begin{table}[ht]
\centering
\small
\resizebox{\columnwidth}{!}{%
\begin{tabular}{c|cc|c|c}
\toprule
 & \multicolumn{2}{c|}{\textbf{RefCOCO+}} & \textbf{RefCOCOg} & \multirow{2}{*}{Avg} \\
Method & TestA & TestB & Test &  \\
\midrule
Real & 23.6\% & 17.7\% & 24.9\% & 22.1\% \\
Ours-top1 & \textbf{31.4\%} & \textbf{21.3\%} & \textbf{30.0\%} & \textbf{27.6\%} \\
Ours-top3 & 30.9\% & 21.1\% & 28.9\% & 27.0\% \\
Ours-top5 & 29.8\% & 20.4\% & 28.3\% & 26.2\% \\
\bottomrule
\end{tabular}%
}
\caption{Performance of different filtering weight configurations on RefCOCO+ and RefCOCOg. Ours-top$k$ refers to the best $k$ filtering-weight configurations selected on the validation set.}
\label{tab:hyperparameter_results}
\end{table}

\subsection{Cross-modality Generative Models}

We evaluate whether POBF can seamlessly integrate with other state-of-the-art pretrained generative models. By default, POBF utilizes Stable Diffusion XL \cite{podellsdxl} as the image generator and BLIP ViT-L \cite{li2022blip} as the image captioner. In this experiment, we test two alternative image generators: Stable Diffusion 2 \cite{rombach2022high}, Dreamshaper\footnote{https://huggingface.co/Lykon/dreamshaper-8-inpainting\label{Dreamshaper}}, and Stable Diffusion 3.5 \cite{rombach2022high}. For image captioning, we explore replacements such as LLAVA \cite{liu2024visual}, ViT-GPT2\footnote{https://huggingface.co/nlpconnect/vit-gpt2-image-captioning\label{ViTGPT2}}, and LLaVA-NeXT \cite{li2024llavanext}. Table \ref{tab:generative_result} shows the results. We observe that while the default configuration of POBF demonstrates strong performance, it does not always yield the highest accuracy. Specifically, model combinations using Stable Diffusion 3.5 as the image generator achieve superior outcomes, reaching an average accuracy of 27.99\%. Furthermore, these configurations consistently outperform the baseline, which highlights the generalizability and effectiveness of our proposed framework.

\begin{table}[ht]
\centering
\small
\resizebox{\columnwidth}{!}{%
\begin{tabular}{c | c c | c | l}
    \toprule
    & \multicolumn{2}{c}{\textbf{RefCOCO+}} & \textbf{RefCOCOg} & \multicolumn{1}{c}{\multirow{2}*{\centering Avg}}\\
    Method & TestA & TestB & Test & \\
    \midrule
    Real & 23.58 & 17.70 & 24.87 & 22.05 \\
    POBF (Ours) & \underline{31.37} & 21.25 & 29.98 & 27.53\textsubscript{\textcolor{green}{+5.48}} \\ 
    \midrule
    \multicolumn{5}{c}{\textit{Replace the Image Captioner with $\dots$}} \\
    LLAVA \cite{liu2024visual} & 31.09 & 21.54 & 28.57 & 27.07\textsubscript{\textcolor{green}{+5.02}} \\ 
    ViT-GPT2 \footref{ViTGPT2} & 29.95 & 21.22 & 29.90 & 27.02\textsubscript{\textcolor{green}{+4.97}} \\ 
    LLaVA-NeXT \cite{li2024llavanext} & 31.15 & 21.62 & 29.75 & \underline{27.51\textsubscript{\textcolor{green}{+5.46}}} \\
    \midrule
    \multicolumn{5}{c}{\textit{Replace the Image Generator with $\dots$}} \\
    SD 2 \cite{rombach2022high} & 31.16 & 21.19 & 29.30 & 27.22\textsubscript{\textcolor{green}{+5.17}} \\ 
    Dreamshaper \footref{Dreamshaper} & 30.20 & \textbf{22.39} & \underline{30.32} & 27.63\textsubscript{\textcolor{green}{+5.58}} \\ 
    SD 3.5 \cite{rombach2022high} & \textbf{31.46} & \underline{21.98} & \textbf{30.53} & \textbf{27.99\textsubscript{\textcolor{green}{+5.94}}} \\
    \bottomrule
\end{tabular}}
\caption{Performance comparison across generative models.}
\label{tab:generative_result}
\end{table}

\begin{table}[ht]
\centering
\small
\resizebox{\columnwidth}{!}{
\begin{tabular}{c|c c|c|l}
    \toprule
    & \multicolumn{2}{c}{\textbf{RefCOCO+}} & \textbf{RefCOCOg} & \multicolumn{1}{c}{\multirow{2}*{\centering Avg}} \\
    Method & TestA & TestB & Test & \\
    \midrule
    \multicolumn{5}{c}{\textit{0.5\% Training Data}} \\
    Real & 16.42 & 10.07 & 15.43 & 13.97 \\ 
    POBF (Ours) & \textbf{21.41} & \textbf{15.87} & \textbf{20.28} & \textbf{19.18\textsubscript{\textcolor{green}{+5.22}}} \\
    \midrule
    \multicolumn{5}{c}{\textit{2.0\% Training Data}} \\
    Real & 33.04 & 25.03 & 33.28 & 30.45 \\ 
    POBF (Ours) & \textbf{37.68} & \textbf{29.88} & \textbf{38.97} & \textbf{35.51\textsubscript{\textcolor{green}{+5.06}}} \\ 
    \bottomrule
\end{tabular}}
\caption{Performance comparison across training data sizes.}
\label{tab:ratio_result}
\end{table}

\subsection{Training Data Sizes}

Table~\ref{tab:ratio_result} presents the performance of our proposed framework under varying amounts of training data. We evaluate two scenarios: using 0.5\% and 2.0\% of the original training set, and report the average accuracy over three randomly sampled datasets per setting. With only 0.5\% of the data, the real-data-only baseline achieves 13.97\% accuracy, while our method improves this to 19.18\%, yielding a substantial gain of 5.22\%. At 2.0\% data, our framework further increases accuracy by 5.06\%. These consistent improvements highlight the robustness and effectiveness of our approach in data-scarce environments, demonstrating its potential for real-world applications where labeled data is limited.

\subsection{Model Architectures}

We evaluate the proposed POBF framework across different model architectures to assess its generalizability and effectiveness. Specifically, we experiment with TransVG \cite{deng2021transvg}, Dynamic-MDTRE \cite{shi2023dynamic}, and QRNet \cite{ye2022shifting}, each of which differs in terms of the number of parameters and architectural complexity. Table \ref{tab:Architectures_result} presents the results. Although model performance generally increases with model size, our POBF framework yields substantial improvements across all tested architectures. Notably, Dynamic-MDTRE shows the largest gain, with a 7.03\% increase in average accuracy. QRNet, the most complex model with 273M parameters, also benefits from a 4.41\% improvement. This highlights that our method is not tailored to a specific architecture but offers consistent benefits across diverse model designs.

\begin{table}
\centering
\resizebox{\columnwidth}{!}{
\begin{tabular}{c|c c|c|l}
    \toprule
    & \multicolumn{2}{c}{\textbf{RefCOCO+}} & \textbf{RefCOCOg} & \multicolumn{1}{c}{\multirow{2}*{\centering Avg}}  \\
    Method & TestA & TestB & Test & \\
    \midrule
    \multicolumn{5}{c}{\textit{TransVG \cite{deng2021transvg} (\#Paras = 149M)}} \\
    Real & 23.58 & 17.70 & 24.87 & 22.05 \\ 
    POBF (Ours) & \textbf{31.37} & \textbf{21.25} & \textbf{29.98} & \textbf{27.53\textsubscript{\textcolor{green}{+5.48}}} \\
    \midrule
    \multicolumn{5}{c}{\textit{Dynamic-MDTRE \cite{shi2023dynamic} (\#Paras = 160M)}} \\
    Real & 23.82 & 18.55 & 29.07 & 23.81 \\ 
    POBF (Ours) & \textbf{33.55} & \textbf{23.93} & \textbf{35.03} & \textbf{30.84\textsubscript{\textcolor{green}{+7.03}}} \\ 
    \midrule
    \multicolumn{5}{c}{\textit{QRNet \cite{ye2022shifting} (\#Paras = 273M)}} \\
    Real & 32.32 & 23.82 & 30.30 & 28.81 \\ 
    POBF (Ours) & \textbf{37.29} & \textbf{26.65} & \textbf{35.73} & \textbf{33.22\textsubscript{\textcolor{green}{+4.41}}} \\
    \bottomrule
\end{tabular}}
\caption{Performance comparison across model architectures. \textit{\#Paras} denotes the number of trainable model parameters.}
\label{tab:Architectures_result}
\end{table}

In addition, We added experiments with different teacher backbones while keeping the student model fixed as TransVG. As shown in Table~\ref{tab:architecture}, all teacher-based filtering variants outperform the no-teacher setting. Moreover, the stronger teacher backbones leads to further improvements, suggesting that the proposed filtering strategy is stable across different teacher architectures.

\begin{table}
\centering
\resizebox{\columnwidth}{!}{
\begin{tabular}{c|c c|c|c}
    \toprule
    & \multicolumn{2}{c}{\textbf{RefCOCO+}} & \textbf{RefCOCOg} & \multicolumn{1}{c}{\multirow{2}*{\centering Avg}} \\
    Teacher & TestA & TestB & Test & \\
    \midrule
    No & 29.10 & 20.10 & 28.31 & 25.83 \\
    TransVG & 31.37 & \underline{21.25} & 29.98 & 27.53 \\
    QRNet & \underline{31.65} & 21.18 & \underline{30.42} & \underline{27.75} \\
    Dynamic-MDTRE & \textbf{32.17} & \textbf{22.08} & \textbf{30.97} & \textbf{28.41} \\
    \bottomrule
\end{tabular}}
\caption{Robustness analysis with different teacher architectures. The student model is fixed as TransVG.}
\label{tab:architecture}
\end{table}

\subsection{Analysis of Bounding-box Size}

We added a stratified analysis based on the target bounding-box size. For each dataset, we sort the test samples by the relative bounding-box area and divide them into three equal groups: the smallest one-third as SMALL, the middle one-third as MEDIUM, and the largest one-third as LARGE. As shown in Table~\ref{tab:size}, POBF improves over the real-only baseline across all groups, but the gain is smaller for SMALL objects. This is likely because training samples with small target boxes require larger outside-box regions to be inpainted, making the generated backgrounds harder to control and more prone to artifacts or noisy contextual cues. Such noise may weaken the model's ability to learn accurate localization for small bounding boxes.

\begin{table*}[h]
\centering
\caption{Performance stratified by target bounding-box size. For each dataset, samples are divided into SMALL, MEDIUM, and LARGE groups according to the relative bounding-box area.}
\label{tab:size}
\small
\setlength{\tabcolsep}{3.5pt}
\begin{tabular}{lccccccccc}
\toprule
\multirow{2}{*}{Method} 
& \multicolumn{3}{c}{RefCOCO TestA} 
& \multicolumn{3}{c}{RefCOCO TestB} 
& \multicolumn{3}{c}{RefCOCOg} \\
\cmidrule(lr){2-4} \cmidrule(lr){5-7} \cmidrule(lr){8-10}
& SMALL & MEDIUM & LARGE 
& SMALL & MEDIUM & LARGE 
& SMALL & MEDIUM & LARGE \\
\midrule
Real-only 
& 40.23 & 44.57 & 47.98 
& 30.57 & 34.61 & 37.84 
& 21.71 & 24.82 & 28.08 \\
POBF      
& 46.52 & 53.89 & 57.90 
& 34.08 & 41.13 & 47.67 
& 25.30 & 29.97 & 34.67 \\
\midrule
Gain      
& +6.29 & +9.32 & +9.92 
& +3.51 & +6.52 & +9.83 
& +3.59 & +5.15 & +6.59 \\
\bottomrule
\end{tabular}
\end{table*}

\subsection{Comparison with Conventional Data Augmentation}

We further compare POBF with two conventional data augmentation methods, Color Jitter and Random Erasing, under the same data-scarce setting. As shown in Table \ref{tab:conventional_augmentation}, Color Jitter and Random Erasing provide only modest average improvements of 0.57\% and 0.25\%, respectively, over training with real data alone. In contrast, POBF improves the average accuracy from 27.78\% to 33.61\% (+5.83\%), demonstrating that synthesizing diverse visual contexts provides substantially greater benefits than conventional image-level augmentation.

\begin{table*}[h]
\centering
\caption{ Comparison with conventional data augmentation methods under the data-scarce visual grounding setting.}
\small
\label{tab:conventional_augmentation}
\begin{tabular}{l|cc|cc|c|c|c}
\toprule
\multirow{2}{*}{Method}
& \multicolumn{2}{c|}{RefCOCO}
& \multicolumn{2}{c|}{RefCOCO+}
& RefCOCOg
& ReferIt
& \multirow{2}{*}{Avg} \\
& TestA & TestB
& TestA & TestB
& Test
& Test
& \\
\midrule

Real
& 44.26 & 34.34
& 23.58 & 17.70
& 24.87 & 21.87
& 27.78 \\

Color Jitter
& 45.12 & 34.71
& 24.75 & 18.03
& 25.11 & 22.36
& 28.35 {\color{green}{$_{+0.57}$}} \\

Random Erasing
& 44.82 & 34.13
& 24.22 & 17.96
& 24.94 & 22.13
& 28.03 {\color{green}{$_{+0.25}$}} \\

POBF (Ours)
& \textbf{52.77} & \textbf{40.96}
& \textbf{31.37} & \textbf{21.25}
& \textbf{29.98} & \textbf{25.28}
& \textbf{33.61} {\color{green}{$_{+5.83}$}} \\

\bottomrule
\end{tabular}
\end{table*}

\subsection{Case Study}

\begin{figure}[t]
  \centering
   \includegraphics[width=1.0\linewidth]{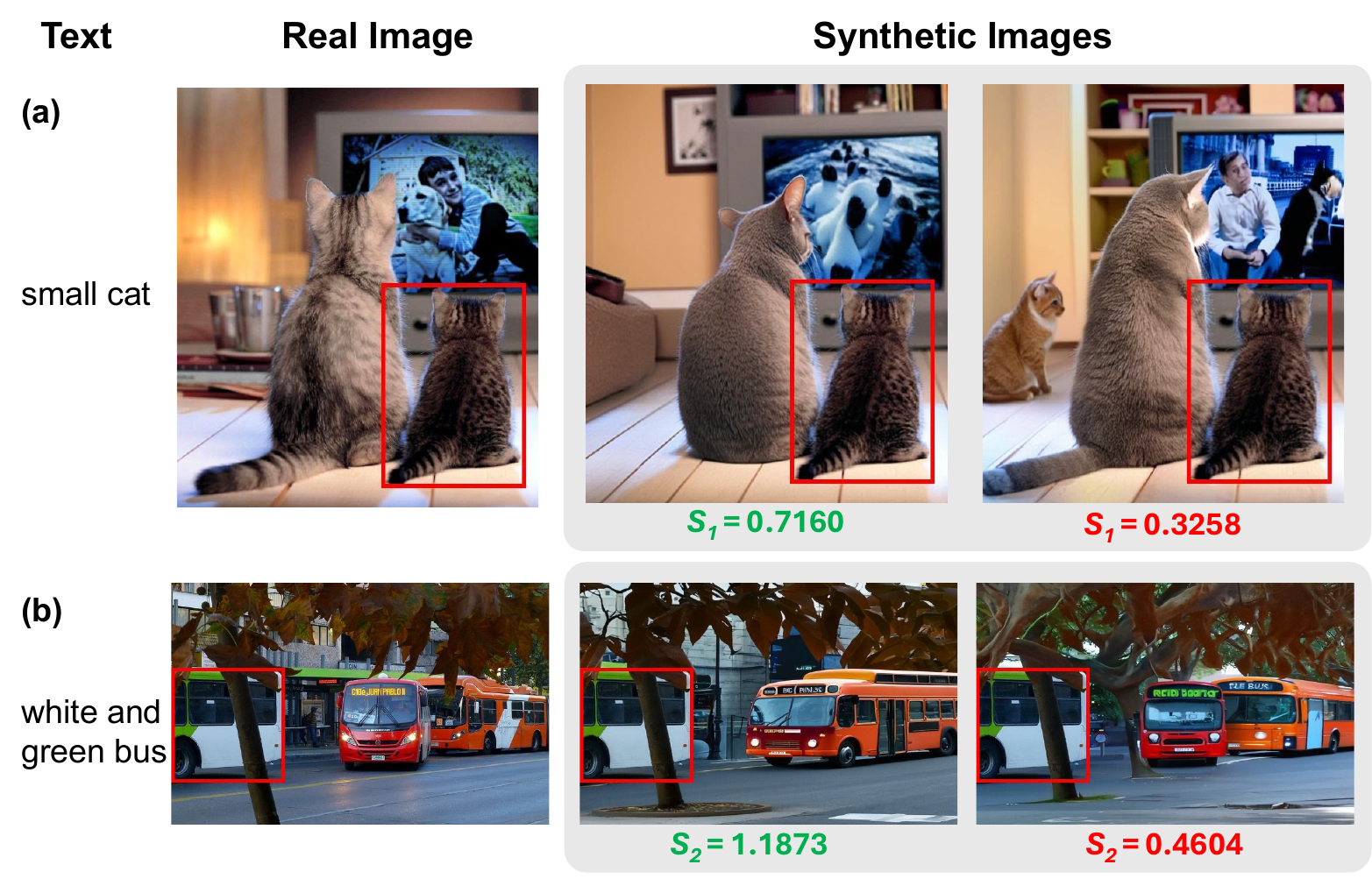}
   \caption{Qualitative examples illustrating the effectiveness of the two proposed scores. The green score indicates the synthetic image with the higher corresponding score.}
   \label{fig:cases}
\end{figure}

Generative models may occasionally produce ambiguous examples. They can also generate synthetic images that closely resemble real ones, increasing the risk of overfitting. In Figure~\ref{fig:cases}, we show two examples illustrating how the hardness ($S_1$) and overfitting ($S_2$) scores help to mitigate these two issues and select the effective data.

In Figure~\ref{fig:cases}(a), the second synthetic image contains a small cat in the background, which may confuse the model when attempting to localize the target object \texttt{small cat}. Since this example is difficult for the teacher model to localize correctly, it receives a low $S_1$ and could thus be filtered out. In Figure~\ref{fig:cases}(b), the background of the second generated image is visually very similar to the real image — the color, position, and posture of the two buses closely resembles those in the real image. Including such a similar sample in training data may increase the risk of overfitting. In contrast, the first generated image background presents a bus with a noticeably different appearance and placement compared to the real image. Although \(S_2\) does not directly measure image similarity, the visually more similar second image receives a lower \(S_2\), which is consistent with our intuition that reproducing training-specific visual patterns may increase the risk of overfitting.

\section{Conclusion}
\label{sec:conclusions}

In this paper, we introduced POBF, a novel framework for effectively learning visual grounding in data-scarce scenarios. Specifically, we proposed a new data generation strategy—paint outside the box—which addresses the label misalignment issues commonly encountered in previous approaches. Additionally, to select the most effective synthetic samples, we introduced a novel filtering scheme tailored for inpainted data. This scheme integrates a hardness score, an overfitting score, and a penalty term. Experimental results demonstrate that POBF surpasses leading baselines, with accuracy gains ranging from 2.29\%–3.85\%. We further validated the robustness and generalizability of POBF across different generative models, training data sizes, and model architectures.

\section*{Acknowledgment}

This research is supported, in part, by the RIE2025 Industry Alignment Fund – Industry Collaboration Projects (IAF-ICP) (Award I2301E0026), administered by A*STAR, as well as supported by Alibaba Group and NTU Singapore through Alibaba-NTU Global e-Sustainability CorpLab (ANGEL). The research is also partially funded by National Research Foundation Fellowship (NRFF13-2021-0006) and MOE AcRF Tier-1 (RG21/26) of Singapore.

\bibliography{sn-bibliography}

\end{document}